\documentclass[10pt,twocolumn,letterpaper]{article}

\usepackage{cvpr}              

\usepackage[dvipsnames]{xcolor}

\definecolor{cvprblue}{rgb}{0.21,0.49,0.74}
\usepackage[pagebackref,breaklinks,colorlinks,citecolor=cvprblue]{hyperref}

\usepackage[capitalize]{cleveref}
\crefname{section}{Sec.}{Secs.}
\Crefname{section}{Section}{Sections}
\Crefname{table}{Table}{Tables}
\crefname{table}{Tab.}{Tabs.}

\def\paperID{11069} 
\def\confName{CVPR}
\def\confYear{2024}

\usepackage{adjustbox}
\usepackage{threeparttable}
\usepackage{xcolor}
\usepackage{multirow}
\usepackage{makecell}
\usepackage{siunitx}
\usepackage{tcolorbox}
\usepackage[accsupp]{axessibility} 
\usepackage{microtype}
\usepackage{marvosym}
\newcommand{\ourmodel}{{\sc {OmniParser}}\xspace}
\newcommand{\pointsdecoder}{{Structured Points Decoder}\xspace}
\newcommand{\polydecoder}{{Region Decoder}\xspace}
\newcommand{\contentdecoder}{{Content Decoder}\xspace}

\newcommand\mypara[1]{\vspace{1.0mm}\noindent\textbf{#1}}
\newcommand\rankfirst[1]{\textbf{#1}}
\newcommand\ranksecond[1]{\underline{#1}}
\newcommand\token[1]{\texttt{\textless{#1}\textgreater}}

\title{\ourmodel: A Unified Framework for Text Spotting, Key Information Extraction and Table Recognition}

\author{
Jianqiang Wan$^1$$^*$~~
Sibo Song$^1$$^*$~~
Wenwen Yu$^2$$^*$~~
Yuliang Liu$^2$\textsuperscript{\Letter} ~~
Wenqing Cheng$^2$~~
\\
Fei Huang$^1$~~
Xiang Bai$^2$~~
Cong Yao$^1$~~
Zhibo Yang$^1$\textsuperscript{\Letter} ~~
\smallskip
\\
$^1$Alibaba Group
\quad
$^2$Huazhong University of Science and Technology
\smallskip
\\ 
\small{\texttt{\{hustwjq,sibosongzju,yangzhibo450,yaocong2010\}@gmail.com}} \\
\small{\texttt{f.huang@alibaba-inc.com}~~~~}
\small{\texttt{\{wenwenyu,ylliu,xbai,chengwq\}@hust.edu.cn}}
}

\begin{document}
\maketitle
\def\thefootnote{*}\footnotetext{Equal contribution. \textsuperscript{\Letter}Corresponding authors.}
\def\thefootnote{\arabic{footnote}}
\begin{abstract}
Recently, visually-situated text parsing (VsTP) has experienced notable advancements, driven by the increasing demand for automated document understanding and the emergence of Generative Large Language Models (LLMs) capable of processing document-based questions. Various methods have been proposed to address the challenging problem of VsTP. However, due to the diversified targets and heterogeneous schemas, previous works usually design task-specific architectures and objectives for individual tasks, which inadvertently leads to modal isolation and complex workflow. In this paper, we propose a unified paradigm for parsing visually-situated text across diverse scenarios. Specifically, we devise a universal model, called OmniParser, which can simultaneously handle three typical visually-situated text parsing tasks: text spotting, key information extraction, and table recognition. In OmniParser, all tasks share the unified encoder-decoder architecture, the unified objective: \textbf{point-conditioned text generation}, and the unified input\&output representation: \textbf{prompt \& structured sequences}.  Extensive experiments demonstrate that the proposed OmniParser achieves state-of-the-art (SOTA) or highly competitive performances on 7 datasets for the three visually-situated text parsing tasks, despite its unified, concise design. The code is available at \href{https://github.com/AlibabaResearch/AdvancedLiterateMachinery}{AdvancedLiterateMachinery}.
\vspace{-2mm}

\end{abstract}

\section{Introduction}
\label{sec:intro}

\begin{figure}[htbp]
    \centering 
    \centerline{\includegraphics[width=1.0\linewidth]{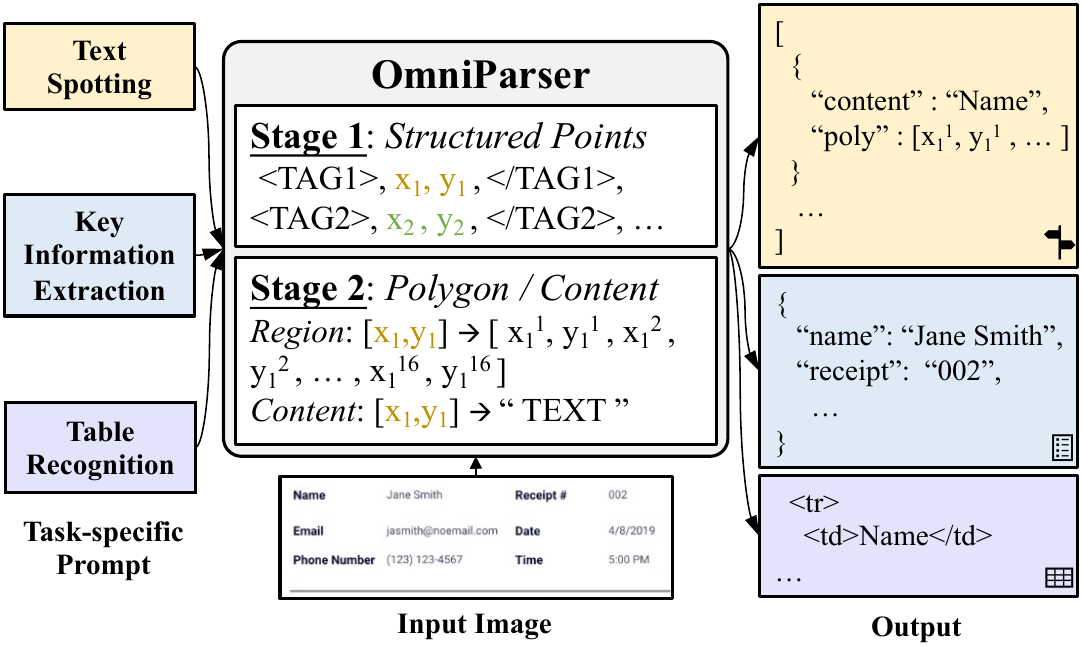}}
    \vspace{-2mm}
    \caption{\textbf{A task-agnostic architecture for visually-situated text parsing. } The proposed \ourmodel takes an image and a task-specific indicator as input and generates structured text sequences tailored to the specified task, including text spotting, key information extraction, and table recognition.}
    \label{fig:pipeline}
    \vspace{-6mm}
\end{figure}

Visually-situated text parsing (VsTP) is designed to extract structured information from document images. It involves the spotting and parsing of textual and visual elements within the text-rich image, such as text, tables, graphics, and other visual entities, partly shown in~\cref{fig:pipeline}. With the rapid growth in the volume of text-related data and the enormous advance in Large Language Models~\cite{ChatGPT, GPT-4} and Multi-modal Large Language Models~\cite{GPT-4V(ision)}, there has been recently a surge of research on the topic of VsTP~\cite{Chen2023PaLIXOS, li2023blip2, li2022relational, ye2023deepsolo}. These methods can be further categorized into generalist models~\cite{Chen2023PaLIXOS, li2023blip2} and specialist models~\cite{ ye2023deepsolo, long2021parsing, wang2021towards}.

Both generalist models and specialist models have limitations in handling multiple multimodal tasks that are closely interconnected in the domain of VsTP. Generalist models excel in their versatility and universality across domains, but fall short in achieving high precision and interpretability. The performances will be restricted if an external OCR engine is not available~\cite{Chen2023PaLIXOS}. Moreover, the prediction processes of such models are usually non-transparent, due to their black-box nature. Regarding specialist models, they frequently achieve higher performance in their respective sub-tasks~\cite{long2021parsing, ye2023deepsolo}. However, when confronted with the requirement of multitasking, the pipeline will be usually more complex. Furthermore, discrete specialist models inadvertently lead to modal isolation and limit in-depth understanding.




In recent years, there has been a trend towards unified models capable of performing multiple visually-situated text parsing tasks, as illustrated in ~\cref{table:unified}. While these models have shown effectiveness, handling diverse text structures and various relations in VsTP remains challenging. Accordingly, tasks in visual document parsing can be categorized into: 1) Sequential text detection and recognition, 2) Table structure and content recognition, and 3) Visual entity extraction and localization. Addressing these diversities while maintaining superior performance in a unified framework poses several challenges. First, incorporating task-specific heads~\cite{ye2023deepsolo}, adapters~\cite{liu2021abcnet, liao2020mask}, and formulations~\cite{long2021parsing, kim2022donut} can hinder achieving generality. Second, handling cross-dependencies between tasks is crucial, for instance, table recognition encompasses text spotting. Third, the unified representation of tasks should consider both primary elements~(\textit{words, points, lines, cells}) and various types of relations~(\textit{the adjacency between characters, the linking between keys and values, and the alignment of table cells.}).  

Along with this line of works, we propose a unified paradigm for visually-situated text parsing in this paper (named \textit{\textbf{OmniParser}}). By adopting a single architecture, standardizing modeling objective as well as output representation, \ourmodel seamlessly handles text spotting, key information extraction (KIE), and table recognition (TR) in a unified framework, as shown in~\cref{fig:pipeline}. To boost performance and increase transparency, we adopt a two-stage generation strategy. In the first stage, a structured sequence consisting of center points of text segments and task-related structural tokens is generated, given the embeddings of the input image and task prompt. In the second stage, polygonal contour and recognition results are predicted for each center point.

The philosophy behind the two-stage design is straightforward. The first stage produces center point sequences which can represent word-level/line-level text instances with complex structures encoded in various markup languages, e.g., JSON or HTML. The second stage can uniformly generate polygonal contours and recognition results across different tasks. An obvious advantage of our two-stage strategy is that the explicit decoupling could greatly reduce the difficulty of learning structured sequences, since the sequence lengths are significantly reduced. As such, higher performance and better generalization ability could be achieved.


To summarize, our major contributions are as follows:
\begin{itemize}
    \item We propose \ourmodel, a unified framework for visually-situated text parsing. To the best of our knowledge, this is the first work that can simultaneously handle text spotting, key information extraction, and table recognition with a single, unified model.
    \item We introduce a two-stage decoder that leverages structured points sequences as an adapter, which not only enhances the parsing capability for structural information, but also provides better interpretability.    
    \item We devise two pre-training strategies, namely spatial-aware prompting and content-aware prompting, which enable a powerful \pointsdecoder for learning complex structures and relations in VsTP. 
    \item Experiments on standard benchmarks demonstrate that the proposed \ourmodel outperforms the existing unified models on the three tasks. Meanwhile, it compares favorably with models with task-specific customization.
\end{itemize}

\begin{table}[t]
\begin{adjustbox}{max width=0.47\textwidth}
\begin{tabular}{lccc}
\toprule
\multirow{2}{*}{Methods}       & \multicolumn{3}{c}{Visually-situated Text Parsing}    \\ \cmidrule(l){2-4} 
                                & Text Spotting & KIE      & Table Recognition    \\ \midrule
Donut~\cite{kim2022donut}      &    $\times$        & E2E, w/o Loc.           &   $\times$        \\
BROS~\cite{hong2022bros}       &     $\times$       & OCR-dependent        & TSR       \\
DocReL~\cite{li2022relational} &     $\times$       & OCR-dependent        & TSR       \\
UniDoc~\cite{feng2023unidoc}   & $\checkmark$        & E2E, w/o Loc.           &  $\times$         \\
SeRum~\cite{cao2023attention}  & $\checkmark$        & E2E, w/o Loc.           &   $\times$        \\ \midrule
\ourmodel                      & $\checkmark$        & E2E           & E2E (TSR + TCR) \\ \bottomrule
\end{tabular}
\end{adjustbox}
\vspace{-2mm}
\caption{\textbf{Comparing the parsing capabilities achieved by different unified paradigms.} `TSR' and `TCR' denote Table Structure Recognition and Table Content Recognition respectively. To the best of our knowledge, \ourmodel is the first paradigm that accomplishes end-to-end visually-situated text parsing for text spotting, key information extraction, and table recognition.}
\label{table:unified}
\vspace{-6mm}
\end{table}

\section{Related Work}
\label{sec:related}

\mypara{Scene Text Spotting.}
Text spotting aims to simultaneously detect and recognize all the texts in an image. Early end-to-end spotting methods~\cite{liu2018fots, li2017towards, he2018end, sun2019textnet, feng2019textdragon}, connected detection and recognition through customized ROI operations, which were not well-suited for curved text. Some segmentation-based methods~\cite{lyu2018mask, qin2019towards, liao2020mask, qiao2021mango} can handle arbitrary-shaped text, but the post-processing and smoothing operations of the segmentation map are not trivial. Recently, transformer-based methods have achieved greater progress with their simple and efficient structures. TESTR~\cite{zhang2022text} utilizes two similar decoders to obtain detection and recognition results separately, while DeepSolo~\cite{ye2023deepsolo} models text semantics and positions explicitly through learnable point queries. However, query-based spotting methods are often limited by the maximum number of detectable texts. Some autoregressive spotting methods can better deal with a large number of texts, such as UNITS~\cite{kil2023towards}, which outputs text sequences using start point prompts until the end. The SPTS series~\cite{peng2022spts, liu2023spts} represent texts with corresponding center points but lack the ability to localize text precisely. 

\mypara{Key Information Extraction.}
Existing KIE approaches can be roughly separated into two categories: OCR-dependent models and OCR-free models. Early research efforts focus on building layout-aware or graph-based representation for KIE via sequence labeling with OCR inputs~\cite{zhou2017east,liao2022real,shi2016end,da2023multi,xu2020layoutlm, xu2021layoutlmv2, huang2022layoutlmv3, xu2021layoutxlm,li2021structurallm,appalaraju2021docformer,li2021selfdoc,yu2021pick,gu2021unidoc,gu2022xylayoutlm,lee2022formnet,peng2022ernie,luo2023geolayoutlm}. However, most of these methods rely on text with proper reading order or extra modules~\cite{wang2021layoutreader, zhang2023reading} for OCR serialization, which is not practical in real-world scenarios.
To address the serialization issue, other methods~\cite{hwang2021spatial,xu2021layoutxlm,hong2022bros,yu2022structextv2,luo2023geolayoutlm,yang2023modeling,zhang2023reading,wei2023ppn} leverage extra detection modules or linking modules for modeling complex relations of text blocks or tokens.
Although these methods employ extra links or modules to solve the reading order issue, the complicated decoding or post-processing strategy limits their generalization ability. 
Beyond that, generation-based methods~\cite{tang2023unifying,cao2023genkie,cao2022query} are proposed to alleviate the burden of post-processing and task-specific link designs. Another category of OCR-free methods employ OCR-aware pre-training or extends with OCR modules in an end-to-end fashion.
Donut and other Seq2Seq-like methods~\cite{kim2022donut,davis2022end,dhouib2023docparser,cao2023attention} adopt a text reading pre-training objective and generate structured outputs consisting of text and entity tokens.
By explicitly equipping text reading modules, previous work~\cite{wang2021towards,tang2021matchvie,zhang2020trie,kuang2023visual,yu2022structextv2} can achieve end-to-end key information extraction with task-specific design.

\mypara{Table Recognition.}
Recent advances in vision-based approaches have improved table extraction from documents, traditionally divided into table detection, table structure recognition (TSR), and table content recognition (TCR). While table detection~\cite{staar2018corpus, zhong2019publaynet} is beyond our scope, TSR, recently adopting an encoder-decoder fashion~\cite{TableMaster, tableformer}, focuses on identifying table structures. TCR involves recognizing text within table cells using established OCR models. Our paper focuses on table recognition (TR), integrating TSR and TCR. TR methods fall into non-end-to-end~\cite{Tsrformer, TRUST, gridformer, VAST} and end-to-end~\cite{EDD, ly2023end} categories. Non-end-to-end methods recover table structure with a specific model and employ offline OCR models for complete HTML sequences. Note that end-to-end table recognition tasks remain less explored due to their complexity and challenging nature.

\mypara{Unified Frameworks.}
We are witnessing a clear trend in building unified frameworks for text-rich image parsing tasks.
Prior arts such as DocReL~\cite{li2022relational} and BROS~\cite{hong2022bros} model relations between table cells or entities through binary classification or a relational matrix, which also requires an off-the-shelf OCR engine.
StrucTexTv2~\cite{yu2022structextv2} proposes a multi-modal learning framework aiming at document image understanding tasks by constructing self-supervised tasks. However, it relies on several task-specific lightweight designs for downstream tasks, such as 
Cascade R-CNN for table cell detection.
Another example, HierText~\cite{long2022towards} pursues unifying scene text detection and layout analysis through an affinity matrix for modeling grouping relations.
Additionally, SeRum~\cite{cao2023attention} converts the end-to-end KIE task into a local decoding process and then shows its effectiveness on text spotting task. 

In this work, we propose \ourmodel that is capable of executing a variety of visually-situated parsing tasks in an end-to-end manner. These tasks encompass text spotting, key information extraction, and table recognition, all of which are consolidated within a unified framework. 
\ourmodel is able to represent the heterogeneous structures of text in natural scenes or document images by decoupling structured points with text regions and contents. 
This bifurcated approach caters to the intrinsic characteristics of text-rich images where the text instances can be parsed concurrently, thereby facilitating an enhancement in universality.

\section{Methodology}

\begin{figure*}[htbp]
    \centering
    \centerline{\includegraphics[width=1.0\linewidth]{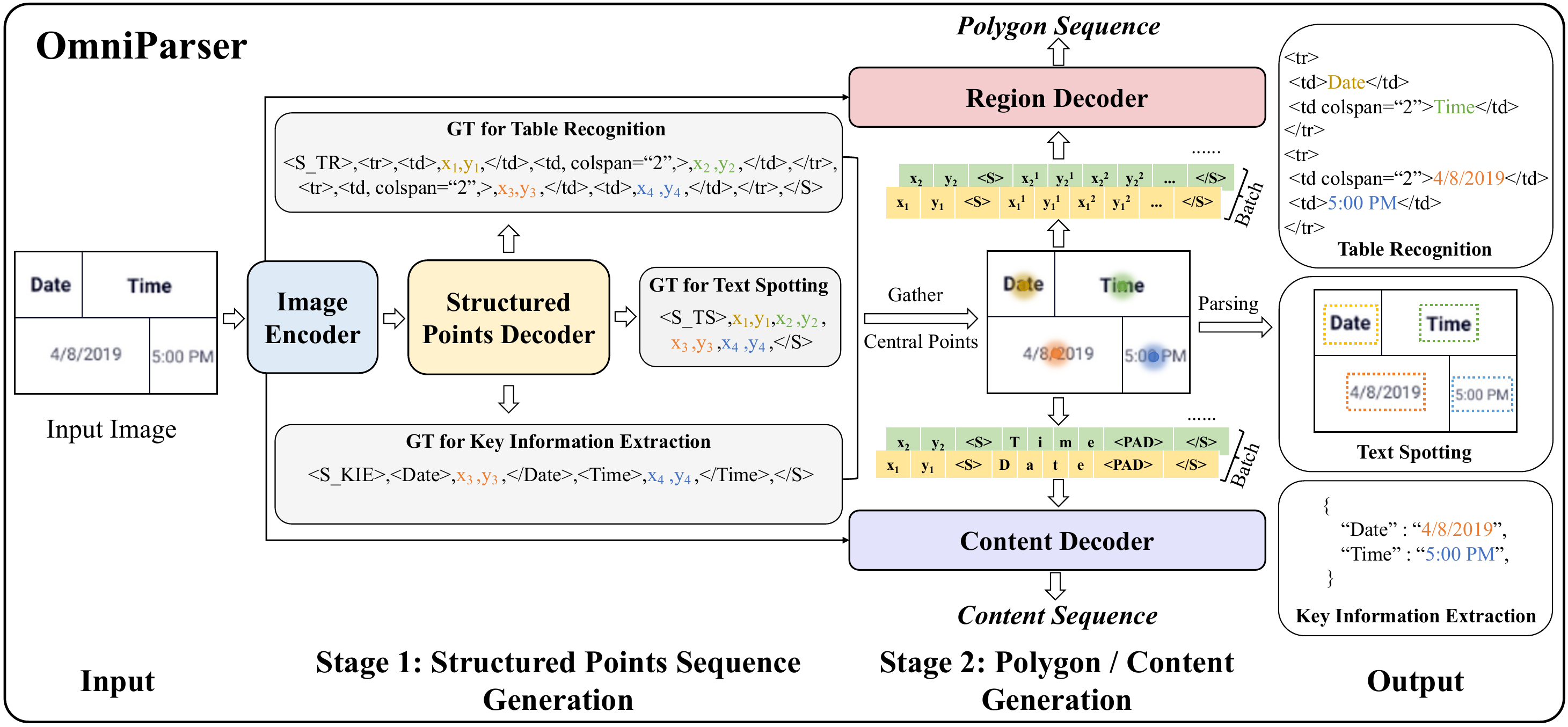}}
    \captionsetup{width=1.0\textwidth}
    \vspace{-2mm}
    \caption{\textbf{Schematic illustration of the proposed OmniParser framework.} \pointsdecoder homogenizes three tasks through a unified structural points representation without designing task-specific branches. 
    Furthermore, benefiting from decoupling points with content recognition and region prediction, the \polydecoder and \contentdecoder can generate polygonal contour and text content in parallel given the text points.}
    \label{fig:diagram}
    \vspace{-6mm}
\end{figure*}


\subsection{Task Unification}
As shown in~\cref{fig:diagram}, we propose a new unified interface that represents structured sequences with three sub-sequences across diverse tasks.
Points are employed as bridges to effectively link structural tags with region and content sequences.

\mypara{Structured Points Sequence Construction}
comprises center points tokens as well as a variety of structural tokens designed for different tasks.
The x and y coordinates of each point are first normalized to the width and height of the image, respectively. Subsequently, they are quantized into discrete tokens within the range of $[0, n_{bins} - 1]$.
Moreover, structural tokens are introduced to represent the entire sequence, such as \token{address} in KIE task and \token{tr} in table recognition task.
Note that text spotting can be seen as a special case that no structural token is incorporated.

\mypara{Polygon \& Content Sequence Construction}
is consistent across all tasks.
We adopt 16-point polygonal formats to represent the polygonal contour for each text instance. 
Each point in the polygon sequence is tokenized following the same procedure as the center point tokenization.
Besides, the transcription of text instances is converted into discrete tokens through char-level tokenization.

\subsection{Unified Architecture}

In light of our overarching goal to enhance the general-purpose paradigm for parsing text-rich images, we utilize a straightforward framework to assess the effectiveness of our proposed representation. To this end, we propose an encoder-decoder architecture that effectively addresses a wide range of visual text parsing tasks, as depicted in~\cref{fig:diagram}.

\mypara{Image Encoder.}
We adopt the Swin-B~\cite{liu2021swin} pre-trained on ImageNet 22k dataset as the fundamental visual feature extractor. 
Specifically, given an image $\mathbf{I} \in \mathbb{R}^{H \times W \times 3}$, we first use the image encoder to extract block-wise visual features which have strides of {4, 8, 16, 32} with respect to the input image. 
Afterward, we employ FPN~\cite{lin2017feature} for feature fusion in order to better capture text features at various scales, following~\cite{song2022vision}. 
Formally, a set of visual embeddings $\left\{\mathbf{v}_i \mid \mathbf{v}_i \in \mathbb{R}^d, 1 \leq i \leq n\right\}$ is generated, where $n$ is feature map size after FPN and $d$ is the dimension of the latent embeddings of the decoders.

\mypara{Decoders.}
\pointsdecoder, \polydecoder, and \contentdecoder are used for structure points sequence generation, detection, and recognition, respectively.
These three decoders share identical network architectures but have independent parameters. 
Each decoder includes four transformer decoder layers with eight heads and pre-attention layer normalization~\cite{xiong2020layer}. 
The hidden dimension of each decoder layer and amplification factor for the MLP layer are set to 512 and 4 respectively. 
Due to varying maximum decoding lengths for the three decoders, we assign uniquely randomly initialized positional encodings to each decoder, aiming to better model the dependencies within the sequences.

\mypara{Objective.}
During pre-training and fine-tuning, the model is trained by minimizing negative log-likelihood given the input sequence $\mathbf{s}$ and visual embeddings $\mathbf{v}$ at $j^{\text{th}}$ time step,

\begin{equation}
L=-\sum_{j=k}^N w_j \log P\left(\mathbf{\tilde{s}}_j \mid \mathbf{v}, \mathbf{s}_{k: j-1}\right) \,,
\label{eq:loss}
\end{equation}
where $\mathbf{\tilde{s}}$ denote the target sequence and $N$ is the length of the sequence. 
Additionally, $w_j$ is the weight value for the $j^{\text{th}}$ token.
We empirically set $w$ to $4.0$ for structural or entity tags and $1.0$ for other tokens.
First $k$ prompt tokens are excluded from the loss calculation.

\subsection{Pre-training Methods}

In our framework, generating structural points sequence is more challenging as it requires \pointsdecoder to understand the text structure and reason entity semantics with image-based input only. 
Therefore, we adopt spatial-aware and content-aware pre-training strategies: spatial-window prompting and prefix-window prompting, to enhance richer spatial and semantic representation learning. 

\begin{figure}[htbp]
    \centering 
    \centerline{\includegraphics[width=1.0\linewidth]{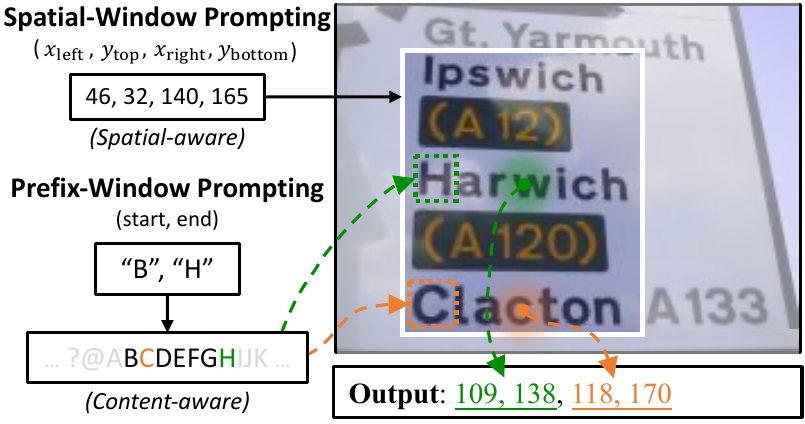}}
    \vspace{-2mm}
    \caption{\textbf{Spatial-Window Prompting} utilizes a 2-point prompt denoted as $(x_{\texttt{left}}, y_{\texttt{top}}, x_{\texttt{right}}, y_{\texttt{bottom}})$, which specifies the location of the prompting spatial window. \textbf{Prefix-Window Prompting} employs a 2-character prompt which indicates the starting and ending characters of the prefix-window with the entire dictionary. The selected prefix range is highlighted in \textbf{black}, while others are shaded in \textcolor{gray}{gray}. The outputs comprise the center points of two words: \text{``Harwich''} and \text{``Clacton''}, as the prefixes `H' and `C' fall within the predefined prefix range.}
    \label{fig:prompting}
    \vspace{-7mm}
\end{figure}

\mypara{Spatial-Window Prompting} guides the \pointsdecoder to read text inside a specified window.
As shown in~\cref{fig:prompting}, only the text center point located in the specified window is considered during training. 
The spatial-window prompting mechanism consists of two patterns: fixed pattern and random pattern.
In the fixed pattern, the window is uniformly sampled from a list of pre-defined layouts, such as $3\times3$ or $2\times2$ grids.
In the random pattern, the window is randomly sampled from an image, ensuring it covers at least $1/9$ of the image. 
More details are provided in the supplementary material.
Similar to Starting-Point Prompting~\cite{kil2023towards}, this spatial-aware prompting strategy allows detecting numerous text from images, even with a limited decoder length.

\mypara{Prefix-Window Prompting} guides the \pointsdecoder to output center points of text with a specified single char prefix.
This strategy aims to instruct the model in locating text instances whose single-character prefix falls within the designated prefix-window charset, while disregarding instances with prefixes outside this charset. The prefix-window charset is sampled from an ordered list of character dictionaries, including 26 uppercase letters, 26 non-capital lowercase, 10 digits, and 34 ASCII punctuation marks, defined by the starting and ending characters.
With the aid of prefix-window prompting, the \pointsdecoder can encode character-level semantics and thus achieve better performance for predicting complex text structures from various tasks such as KIE.

\section{Experiments}

In this section, we conduct both qualitative and quantitative experiments on standard benchmarks, to verify the effectiveness and advantages of the proposed \ourmodel.

\begin{table*}[htpb]
\resizebox{\textwidth}{!}{
\begin{tabular}{lcccccccccccccccc}
\toprule
\multirow{3}{*}{Methods}             & \multicolumn{5}{c}{Total-Text}   & \multicolumn{5}{c}{CTW1500}      & \multicolumn{6}{c}{ICDAR 2015}          \\ \cmidrule(lr){2-6} \cmidrule(lr){7-11} \cmidrule(lr){12-17}
& \multicolumn{3}{c}{Detection}   & \multicolumn{2}{c}{E2E}    & \multicolumn{3}{c}{Detection}   & \multicolumn{2}{c}{E2E} & \multicolumn{3}{c}{Detection}   & \multicolumn{3}{c}{E2E}          \\
\cmidrule(lr){2-4} \cmidrule(lr){5-6} \cmidrule(lr){7-9}  \cmidrule(lr){10-11} \cmidrule(lr){12-14} \cmidrule(lr){15-17} 
                    & P    & R    & F    & None & Full & P    & R    & F    & None & Full & P    & R    & F    & S    & W    & G  
                    \\ \midrule
TextDragon~\cite{feng2019textdragon}          & 85.6 & 75.7 & 80.3 & 48.8 & 74.8 & 82.8 & 84.5 & 83.6 & 39.7 & 72.4 & 92.5 & 83.8 & 87.9 & 82.5 & 78.3 & 65.2 \\
CharNet~\cite{xing2019convolutional}             & 88.6 & 81.0 & 84.6 & 63.6 & -    & -    & -    & -    & -    & -    & 91.2 & 88.3 & 89.7 & 80.1 & 74.5 & 62.2 \\
TextPerceptron~\cite{qiao2020text}      & 88.8 & 81.8 & 85.2 & 69.7 & 78.3 & -    & -    & -    & 57.0 & -    & 92.3 & 82.5 & 87.1 & 80.5 & 76.6 & 65.1 \\
CRAFTS~\cite{baek2020character}              & 89.5 & 85.4 & 87.4 & 78.7 & -    & -    & -    & -    & -    & -    & 89.0 & 85.3 & 87.1 & 83.1 & 82.1 & 74.9 \\
Boundary~\cite{wang2020all}            & 88.9 & 85.0 & 87.0 & 65.0 & 76.1 & -    & -    & -    & -    & -    & 89.8 & 87.5 & 88.6 & 79.7 & 75.2 & 64.1 \\
Mask TextSpotter v3~\cite{liao2020mask} & -    & -    & -    & 71.2 & 78.4 & -    & -    & -    & -    & -    & -    & -    & -    & 83.3 & 78.1 & 74.2 \\
PGNet~\cite{wang2021pgnet}               & 85.5 & 86.8 & 86.1 & 63.1 & -    & -    & -    & -    & -    & -    & 91.8 & 84.8 & 88.2 & 83.3 & 78.3 & 63.5 \\
MANGO~\cite{qiao2021mango}               & -    & -    & -    & 72.9 & 83.6 & -    & -    & -    & 58.9 & 78.7 & -    & -    & -    & 85.4 & 80.1 & 73.9 \\
PAN++~\cite{wang2021pan++}               & -    & -    & -    & 68.6 & 78.6 & 87.1 & 81.0   & 84.0 & -    & -    & -    & -    & -    & 82.7 & 78.2 & 69.2 \\
ABCNet v2~\cite{liu2021abcnet}           & 90.2 & 84.1 & 87.0 & 70.4 & 78.1 & 83.8 & 85.6 & 84.7 & 57.5 & 77.2 & 90.4 & 86.0 & 88.1 & 82.7 & 78.5 & 73.0 \\
TPSNet~\cite{wang2022tpsnet}              & 90.2 & 86.8 & 88.5 & 76.1 & 82.3 & -    & -    & -    & 59.7 & 79.2 & -    & -    & -    & -    & -    & -    \\
ABINet++~\cite{fang2022abinet++}            & -    & -    & -    & 77.6 & 84.5 & -    & -    & -    & 60.2 & 80.3 & -    & -    & -    & 84.1 & 80.4 & 75.4 \\ 
GLASS~\cite{ronen2022glass}               & 90.8 & 85.5 & 88.1 & 79.9 & 86.2 & -    & -    & -    & -    & -    & 86.9 & 84.5 & 85.7 & 84.7 & 80.1 & 76.3 \\
TESTR~\cite{zhang2022text}               & 93.4 & 81.4 & 86.9 & 73.3 & 83.9 & 92.0 & 82.6 & 87.1 & 56.0 & 81.5 & 90.3 & 89.7 & 90.0 & 85.2 & 79.4 & 73.6 \\
SwinTextSpotter~\cite{huang2022swintextspotter}    & -    & -    & 88.0 & 74.3 & 84.1 & -    & -    & 88.0 & 51.8 & 77.0 & -    & -    & -    & 83.9 & 77.3 & 70.5 \\
SPTS~\cite{peng2022spts}                & -    & -    & -    & 74.2 & 82.4 & -    & -    & -    & 63.6 & \ranksecond{83.8} & -    & -    & -    & 77.5 & 70.2 & 65.8 \\
TTS~\cite{kittenplon2022towards}                 & -    & -    & -    & 78.2 & 86.3 & -    & -    & -    & -    & -    & -    & -    & -    & 85.2 & 81.7 & 77.4 \\
UNITS~\cite{kil2023towards}                 & -    & -    & 89.8    & 82.2 & 88.0 & -    & -    & 88.6    & \ranksecond{66.4}    & 82.3    & 91.0    & 94.0    & 92.5    & \ranksecond{89.0} & 84.1 & \rankfirst{80.3} \\
DeepSolo~\cite{ye2023deepsolo}            & 93.2 & 84.6 & 88.7 & \ranksecond{82.5} & \ranksecond{88.7} & -    & -    & -    & 56.7 & -    & 92.5 & 87.2 & 89.8 & 88.0 & 83.5 & 79.1 \\
DeepSolo$^*$~\cite{ye2023deepsolo}            & 92.8  & 82.4  & 87.4 & 81.2 & 87.8 & 91.5    & 84.8   & 88.0    & 64.9 & 81.2    & 92.4 & 88.8 & 90.6 & 88.9 & \ranksecond{84.4}  & 79.5  \\
\midrule
\ourmodel (ours)                & 88.4 & 88.6 & 88.5 & \rankfirst{84.0} & \rankfirst{88.9} & 87.9 & 87.6 & 87.8 & \rankfirst{66.8} & \rankfirst{85.1} & 90.3 & 91.0 & 90.7 & \rankfirst{89.6} & \rankfirst{84.5} & \ranksecond{79.9} \\
\bottomrule
\end{tabular}
}
\vspace{-2mm}
\caption{\textbf{Comparisons on text spotting task.} `S', `W', and `G' refer to the spotting performance obtained by utilizing strong, weak, and generic lexicons, respectively. The end-to-end metrics are highlighted as they are the primary metrics for text spotting. Bold and underline denote the first and second performances, respectively. $^*$ indicates the use of open-source code on our dataset configuration.
}
\label{table:textspotting}
\vspace{-4mm}
\end{table*}

\subsection{Implementation Details}
\mypara{Pre-training.}
\ourmodel is first trained on a hybrid dataset containing Curved SynthText~\cite{liu2021abcnet}, ICDAR 2013~\cite{karatzas2013icdar}, ICDAR 2015~\cite{karatzas2015icdar}, MLT 2017~\cite{nayef2017icdar2017}, Total-Text~\cite{ch2020total}, TextOCR~\cite{singh2021textocr}, HierText~\cite{long2022towards}, COCO Text~\cite{gomez2017icdar2017}, and Open Image V5~\cite{krylov2021open}. 
To accelerate convergence, we adopt a two-stage pre-training strategy following Pix2seq~\cite{chen2021pix2seq}. In the first stage, the model is trained with a batch size of 128 and image resolution of $768\times768$ for 500k steps. Subsequently, we continue training for an additional 200k steps with a batch size of 16 and image resolution of $1920\times1920$. Both stages utilize the AdamW~\cite{loshchilov2018decoupled} optimizer, with initial learning rates of \num{5e-4} and \num{2.5e-4}, respectively. Warm-up schedule is used for the first 5k steps, after which the learning rate is linearly decayed to 0. For data augmentation, we employ instance-aware random cropping, random rotation between $-90^\circ$ and $90^\circ$, random resizing, and color jittering. 
During pre-training, the center points of text instances are arranged in a raster scan order.

\mypara{Fine-Tuning.}
For text spotting and KIE tasks, the model is fine-tuned on the corresponding dataset for 20k and 200k steps respectively, with a learning rate set to \num{1e-4}. 
For table recognition, the default maximum sequence lengths for \pointsdecoder and \contentdecoder are set to 1,500 and 200, respectively. 
The \pointsdecoder is trained for 400k steps and the \contentdecoder is trained for 200k steps with the learning rate set to \num{1e-4}.
For all tasks, the cosine learning rate scheduler is utilized.
Besides, the spatial-window prompting and prefix-window prompting are modified as $[0, 0, n_{bins} - 1, n_{bins} - 1]$ and [${\texttt{char}_\texttt{first}}$, ${\texttt{char}_\texttt{last}}$] (`!' and `\texttt{\texttildelow}' in the dictionary) respectively, to cover full spatial and prefix range. 


\subsubsection{Text Spotting}

\mypara{Datasets.}
We conduct experiments on three popular scene text datasets, Total-Text, ICDAR 2015, and CTW1500~\cite{liu2019curved}. 
Total-Text is mainly for arbitrary-shaped text detection and spotting evaluation, consisting of 1255 training images and 300 testing images with word-level polygon annotations. 
The ICDAR 2015 dataset contains 1000 training images and 500 testing images, annotated with quadrilateral bounding boxes. 
CTW1500 is another benchmark for curved text detection and recognition, which is annotated at text-line level, including 1000 training images and 500 testing images.

\mypara{Evaluation Metrics.}
For Total-Text and CTW1500, we report the end-to-end recognition results over two lexicons: ``None'' and ``Full''. ``None'' means that no lexicons are provided, and ``Full'' lexicon provides all words in the test set.
For ICDAR 2015, we report results over three lexicons: ``Strong'', ``Weak'' and ``Generic''. Strong lexicon provides 100 words that may appear in each image. Weak lexicon provides words in the whole test set, and generic lexicon provides a 90k vocabulary.

\subsubsection{Key Information Extraction}

\mypara{Datasets.}
We evaluate our model's performance on two commonly used benchmark datasets for KIE task: CORD~\cite{park2019cord} and SROIE~\cite{huang2019icdar2019}. 
CORD~\cite{park2019cord} consists of 30 labels across 4 categories.
It has 1,000 receipt samples. The train, validation, and test splits contain 800, 100, and 100 samples respectively. 
The SROIE dataset~\cite{huang2019icdar2019} comprises a training set with 626 receipts and a test set with 347 receipts. Each receipt in the dataset contains four predefined entities, namely: ``company'', ``date'', ``address'', and ``total''. Annotations in the dataset provide segment-level bounding boxes for the text regions and their corresponding transcriptions. 

\mypara{Evaluation Metrics.}
Following~\cite{kim2022donut}, two evaluation metrics are used to evaluate the performance: field-level F1 measure and tree-edit-distance-based accuracy.
The field-level F1 score checks whether each extracted field corresponds exactly to its value in the ground truth.

\subsubsection{Table Recognition}

\mypara{Datasets.} 
Given our model's dual prediction of table logical structures (with cell bounding box central points) and cell content, datasets lacking annotations for both cell content and corresponding bounding boxes, as well as those using metrics incompatible with our approach, are excluded from evaluation. For model assessment, PubTabNet (PTN)~\cite{EDD} and FinTabNet (FTN)~\cite{GTE} are selected. \textbf{PubTabNet} has 500,777 training images and 9,115 validation images, featuring diverse structures from scientific documents. Our model is evaluated on the validation set due to the lack of public annotations for the test set. \textbf{FinTabNet} comprises 112k single-page PDFs with 92,000 cropped training images and 10,656 testing images. 


\mypara{Evaluation Metrics.} For evaluation, we utilized Tree-Edit-Distance-based Similarity (TEDS)~\cite{EDD}. TEDS comprehensively evaluates table similarity, considering both structural and cell content aspects in HTML format. The metric represents the HTML table as a tree, and the TEDS score is computed through the tree-edit distance between the ground truth and predicted trees. In addition to overall results, we also provide S-TEDS results, focusing exclusively on the structural aspects and ignoring cell content.


\subsection{Comparisons with State-of-The-Art}

\mypara{Text Spotting.}
In~\cref{table:textspotting}, we compare \ourmodel with previous text spotting approaches.
On arbitrarily shaped text datasets, Total-Text~\cite{ch2020total} and CTW1500~\cite{liu2019curved}, our method establishes new state-of-the-art under two end-to-end metrics.
In particular, our method surpasses previous SOTA by \textcolor{ForestGreen}{+1.5\%} and \textcolor{ForestGreen}{+3.2\%} on Total-Text and CTW1500 respectively without lexicon, outperforming all the other competitors.
It should be noted that our approach achieves comparable detection results, meanwhile outperforming previous work by a significant margin under the end-to-end metrics. 
We attribute this superior performance to the decoupling of the detection and recognition processes. 
On ICDAR 2015 dataset, our method surpasses other approaches, with the exception of the UNITS on generic setting. 
We presume that joint learning heterogeneous region representations such as bounding boxes, quadrilaterals, and polygons can boost detection performance for tiny and distorted text on the ICDAR 2015, therefore facilitating end-to-end spotting.
However, to ensure a more cohesive and standardized region representation, we adopt a 16-point polygonal representation across various visually-situated text parsing tasks.


\begin{table}[t]
   \centering
   \begin{adjustbox}{max width=0.45\textwidth}
   \begin{threeparttable}
     \centering
     \begin{tabular}{lccccc}
     \toprule
     \multirow{2}{*}{Methods} & \multirowcell{2}{Localization \\ Ability} & \multicolumn{2}{c}{CORD} & \multicolumn{2}{c}{SROIE}  \\
      \cmidrule(lr){3-4} \cmidrule(lr){5-6}
      &  & F1 & Acc & F1 & Acc  \\
      \midrule 
        TRIE~\cite{zhang2020trie}                     & Yes  & -    & -     & 82.1 & -   \\
        Donut~\cite{kim2022donut}                     & No   & 84.1 & \rankfirst{90.9}  & 83.2  & \ranksecond{92.8} \\
        Dessurt~\cite{davis2022end}                   & No   & 82.5 & -     & 84.9  & -    \\
        DocParser~\cite{dhouib2023docparser}          & No   & \ranksecond{84.5} & -     & \rankfirst{87.3} & -    \\
        SeRum~\cite{cao2023attention}                 & No   & 80.5 & 85.8  & \ranksecond{85.6} & \ranksecond{92.8} \\
        \midrule
        \ourmodel (ours)                 & Yes      & \rankfirst{84.8} & \ranksecond{88.0} & \ranksecond{85.6}$^\dagger$ & \rankfirst{93.6}$^\dagger$  \\
        \bottomrule
     \end{tabular}
   \end{threeparttable}
   \end{adjustbox}
   \vspace{-2mm}
   \caption{{\bf Comparisons of end-to-end methods on key information extraction.} `F1' denotes the field-level F1 score and `Acc' denotes the tree-edit-distance-based accuracy. $^\dagger$ Since the SROIE dataset does not provide the necessary point location for each entity word, we generate these locations for evaluation purposes.}
   \label{table:kie}
\vspace{-5mm}
\end{table}

\begin{table}[htbp]
\centering
   \begin{adjustbox}{max width=0.47\textwidth}
   \begin{threeparttable}
\begin{tabular}{lccccc}
\toprule
 \multicolumn{5}{c}{PubTabNet (PTN)}       \\
 \midrule
  Methods                         & Input Size      & Decoder Len.      & S-TEDS & TEDS  \\
 \midrule
WYGIWYS~\cite{deng2017image}                                              & 512            & -                & -   & 78.6  \\
  Donut*~\cite{kim2022donut}                                              & 1,280            & 4,000                &  25.28  & 22.7  \\
  EDD~\cite{EDD}                                          & 512             & 1,800                & 89.9   & 88.3  \\
 \midrule
       \multirow{1}{*}{\ourmodel (ours)}      & 1,024            & 1,500             & \textbf{90.45}  & \textbf{88.83} \\
 
\toprule

 \multicolumn{5}{c}{FinTabNet (FTN)}       \\
 
 \midrule
  Methods                            & Input Size      & Decoder Len.      & S-TEDS & TEDS  \\
 \midrule

  Donut*~\cite{kim2022donut}                                                  & 1,280            & 4,000                &  30.66  & 29.1  \\
  EDD~\cite{EDD}                                                    & 512             & 1,800                & 90.6   & -     \\
 \midrule
\multirow{1}{*}{\ourmodel (ours)}  & 1,024            & 1,500             & \textbf{91.55}  & \textbf{89.75} \\

\bottomrule
\end{tabular}
   \end{threeparttable}
   \end{adjustbox}
   \vspace{-2mm}
\caption{\textbf{Comparisons of end-to-end table recognition methods on PubTabNet and FinTabNet datasets.} * represents our reproduced results, where the model was finetuned on PubTabNet and FinTabNet, respectively.}
\label{tab:ptn_and_ftn}
\vspace{-5mm}
\end{table}

\mypara{Key Information Extraction.}
~\cref{table:kie} reports the performance of KIE task compared to state-of-the-art end-to-end methods on CORD and SROIE datasets. 
We have exclusively reported SeRum\textsubscript{total}~\cite{cao2023attention} since all generation-based methods utilize a schema that encompasses the entire token sequence of all key information, making it directly comparable.
Our model achieves an $84.8\%$ field-level F1 score on CORD, outperforming previous generation-based approaches. 
In addition, our method achieves the best TED-based accuracy on SROIE, indicating its superior character-level prediction performance. Notably, the proposed paradigm ensures accurate localization, which is essential for detailed document analysis and correction, a deficiency of other generation-based approaches.
Moreover, in contrast to prior studies that utilized a massive corpus of document data for pre-training, our model is pre-trained on scene text data only. 
This highlights the exceptional generalizability of our unified model.

\mypara{Table Recognition.}
In~\cref{tab:ptn_and_ftn}, we compare \ourmodel's performance with end-to-end table recognition models. Specifically, we fine-tuned the OCR-free model Donut~\cite{kim2022donut} for table recognition with the official default training configuration. Experimental results show that \ourmodel consistently outperforms previous end-to-end methods in TEDS and S-TEDS on various datasets. It's noteworthy that non-end-to-end table structure recognition models~\cite{TableMaster, tableformer, Tsrformer, TRUST, gridformer, VAST} use bounding boxes of cell contents for model training and employ offline OCR models for constructing final complete HTML sequences. In contrast, \ourmodel utilizes points, achieving comparable results in an end-to-end manner, simplifying post-processing and requiring fewer annotations compared to box-based methods.

\section{Analysis}
In this section, we begin by conducting ablation experiments on crucial designs in \ourmodel. We evaluate these ablations using the Total-Text and ICDAR 2015 text spotting tasks. Furthermore, we provide visualizations on downstream tasks to illustrate the effectiveness of \ourmodel.

\begin{table}[htbp]
\centering
\begin{adjustbox}{max width=0.44\textwidth}
\begin{tabular}{ccccccc}
\toprule
\multicolumn{2}{c}{Window-Prompting}& \multicolumn{2}{c}{Total-Text} & \multicolumn{3}{c}{ICDAR 2015} \\
\cmidrule(lr){1-2} \cmidrule(lr){3-4} \cmidrule(lr){5-7}
Spatial-     & Prefix-    & None           & Full          & S        & W        & G   \\ \midrule
  &  &82.4  &87.6  & 88.1 & 83.0  & 78.3          \\
  &     \checkmark  &82.9  &88.1  &88.4  &83.2          & 78.5         \\
  \checkmark        &  &83.5  &88.5  & 89.2  & 84.2  & 79.4          \\ 
  \checkmark  &  \checkmark  & 84.0  & 88.9          & 89.6     & 84.5     & 79.9    \\
\bottomrule
\end{tabular}
\end{adjustbox}
\vspace{-2mm}
\caption{\textbf{Ablation of pre-training strategies} on text spotting. }
\label{table:abpretraintask}
\vspace{-6mm}
\end{table}

\mypara{Ablating Pre-training Strategies.}
To investigate the effects of spatial-window prompting and prefix-window prompting techniques, we conduct ablative experiments and present the findings in~\cref{table:abpretraintask}.
The inclusion of spatial-window prompting yields a significant enhancement in the performance of our model.
This improvement can be attributed to the heightened perception of spatial coordinate positions, thereby enabling more accurate predictions of structured point sequences.
Similarly, the incorporation of prefix-window prompting also results in a noticeable improvement in performance, as it enhances the model's ability to perceive diverse textual content within images.
The spatial-window prompting and prefix-window prompting enhance the model's perception ability in coordinate space and semantic space respectively. 
Notably, when both prompting techniques are employed simultaneously, the model achieved state-of-the-art performance on both datasets.

\begin{table}[htbp]
\vspace{-2mm}
\centering
\begin{adjustbox}{max width=0.47\textwidth}
\begin{tabular}{ccccccc}
\toprule
\multirowcell{2}{Visual \\ Backbone} & \multirow{2}{*}{Decoder} & \multicolumn{2}{c}{Total-Text} & \multicolumn{3}{c}{ICDAR 2015} \\ 
\cmidrule(lr){3-4} \cmidrule(lr){5-7}
                  &          & None           & Full          & S        & W        & G        \\ \midrule
  ResNet50  & Not Shared &  82.1              &   87.1            & 88.2         & 83.0         & 78.4         \\
  Swin-B      & Shared & 82.5           & 87.3          & 88.5     &  83.2    & 78.7      \\
  Swin-B   & Not Shared   & 84.0           & 88.9          & 89.6 & 84.5 & 79.9     \\
\bottomrule
\end{tabular}
\end{adjustbox}
\vspace{-2mm}
\caption{\textbf{Ablation of encoder and decoder designs} on the text spotting task. }
\label{table:abbackbone}
\vspace{-2mm}
\end{table}

\mypara{Ablating Architectural Designs.}
We conduct a comparative analysis of various architectural designs for both the visual encoder and decoders, as presented in~\cref{table:abbackbone}.
As our model comprises three decoders that share the same architecture, we aim to investigate whether weight sharing among these decoders can enhance the overall performance. 
However, our observations reveal that when employing a shared decoder, the performance on text spotting tasks diminishes, suggesting a potential discrepancy among the subtasks of decoding center points, polygons, and content. Additionally, we compare the backbones of ResNet50 and Swin-B. Remarkably, Swin-B outperforms ResNet50, demonstrating its superiority in visually-situated text parsing tasks.



\begin{figure*}[htbp]
\vspace{-2mm}
  \centering 
  \centerline{\includegraphics[width=1.0\linewidth]{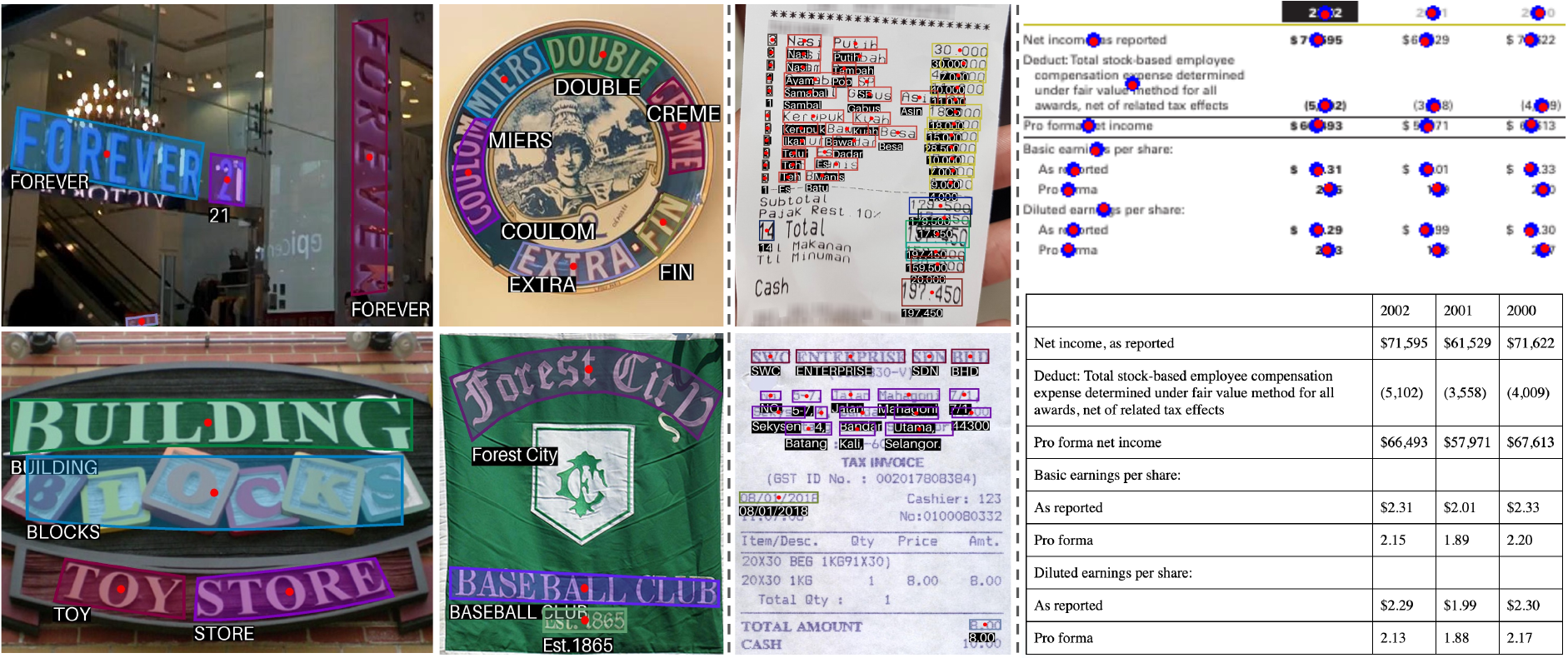}}
  \vspace{-1mm}
  \caption{ \textbf{Qualitative results} of text spotting (column 1-2), KIE (column 3), and table recognition (column 4). For KIE, points, polygons, and recognition are visualized. The color assigned to polygons indicates the entity type. For table recognition, we present point locations and a rendered table based on the prediction sequence, with an additional border for readability. Blue points and red points denote the GT and predicted points respectively. More details can be found in the supplementary material. (The figure is best viewed in color.)}
  \label{fig:visualization}
  \vspace{-4mm}
\end{figure*}

\begin{table}[htbp]
\centering
   \begin{adjustbox}{max width=0.47\textwidth}
   \begin{threeparttable}
\begin{tabular}{lccccc}
\toprule
 \multicolumn{5}{c}{PubTabNet (PTN)}       \\
 \midrule
  Methods                         & S-Decoder Len.      & C-Decoder Len.      & S-TEDS & TEDS  \\
 \midrule
  \multirow{3}{*}{\textbf{\ourmodel}}  & 1,124            & 200             & 89.94  & 88.21 \\

  & 1,500            & 200             & \textbf{90.45} & 88.83 \\
  & 2,000            & 300             & \textbf{90.45}  & \textbf{88.96} \\
\bottomrule
\end{tabular}
   \end{threeparttable}
   \end{adjustbox}
\vspace{-2mm}
\caption{\textbf{Ablation of decoder length for the table recognition task on PubTabNet datasets.} S-Decoder Len. and C-Decoder Len.: short for the length of \pointsdecoder and \contentdecoder, respectively. }
\label{tab:ptn_var_dec_len}
\end{table}

\mypara{Ablating Decoder Length.}
In~\cref{tab:ptn_var_dec_len}, we perform an ablation study on decoder lengths for end-to-end table recognition. Due to GPU constraints, Donut's max length is set to 4,000 (shown in~\cref{tab:ptn_and_ftn}), while our model at 1,500 achieves better results. Note that the average inference speed of our method and Donut are 1.3 and 0.8 FPS, respectively. Training end-to-end models like Donut with complete HTML sequences poses challenges for lengthy sequences, such as those encountered in table recognition, where there is a high probability of error accumulation and attention drift. Our modularized architecture separates pure table HTML tags and cell text sequences, enabling end-to-end recognition without length restrictions. Besides, increasing the length of \pointsdecoder from 1,500 to 2,000 shows no improvement in S-TEDS, with slight TEDS enhancement when the text length increases from 200 to 300. In practice, decoder length choice requires a trade-off between performance and efficiency.

\mypara{Qualitative Results.} We show qualitative results for three tasks in~\cref{fig:visualization}: 1) For text spotting, our model can accurately detect and recognize curve texts, vertical texts, and artistic texts under challenging scenarios. Despite some imprecise detections, the recognition results are entirely accurate. 2) In table recognition results, hard cases of spanning cells, borderless tables, and cells with multi-line content are presented.
These examples show that our method can correctly localize cell centers through the structured points sequence. 3) KIE results demonstrate the efficacy of our approach in effectively localizing, recognizing texts and, more importantly, extracting entity information.



\noindent\textbf{Limitations.} Despite achieving promising results on visually-situated text tasks, the proposed \ourmodel has a few limitations. Firstly, it relies on having precise word point locations during training, which may not be always available in certain real-world scenarios. Secondly, it does not account for parsing non-text elements such as figures or charts, limiting its potential in solving complex document parsing tasks. Addressing such limitations and improving the robustness as well as the applicability of our model in real-world settings will be the focus of our future research.

\section{Conclusions and Future Works}

In this paper, we have proposed a general-purpose parsing framework \ourmodel, which brings together the tasks of text spotting, key information extraction, and table recognition in a visually-situated text parsing context. 
This is realized through a two-stage decoding procedure, leveraging structured points as an adapter. To enhance the effectiveness of pre-training across all tasks, we also introduce two pre-training strategies to enable the \pointsdecoder to learn complex structures and relations among visually-situated texts, further improving the overall performance. 

The proposed \ourmodel achieves state-of-the-art or highly competitive performance on standard benchmarks, even compared with specialist models that rely on task-specific designs. As a general-purpose parser, \ourmodel has been proven quite effective on 
various visually-situated text tasks, so we will extend it to more tasks and scenarios, e.g., layout analysis and chart parsing.

\noindent\textbf{Acknowledgements.} This work was supported by the National Natural Science Foundation of China (No.62225603), and Alibaba Innovative Research (AIR) program.

{
    \small
    \bibliographystyle{ieeenat_fullname}
    \bibliography{main}
}


\end{document}


\maketitle

\section{Implementation Details}
\subsection{Spatial-Window Prompting}
Spatial-window prompting comprises two components: fixed mode and random mode. In the fixed mode, the image is divided into grid blocks evenly, such as 3x3 or 2x2. Conversely, in the random mode, the starting point of spatial window is randomly determined. In order to encompass more texts within the random box, the area of the random box is established to be no less than 1/9 of the original image. To elaborate further, a 30\% probability is assigned for selecting the fixed mode, another 30\% probability for selecting the random mode, and a 40\% probability for defaulting window to cover the entire image. Following ~\cite{kil2023towards}, we set the bin size of coordinate vocab as 1000. The pseudo-code of spatial-window prompting is shown in the following.

\begin{lstlisting}[language=Python]
import random

# prob for different mode
prob = random.uniform(0, 1)

# quantizing coordinates with n_bins
n_bins = 1000

if prob < 0.4:
    # default window
    start_x, start_y, end_x, end_y = [0, 0, n_bins - 1, n_bins - 1]
elif prob < 0.7:
    # x-axis and y-axis are partitioned into varying numbers of blocks.
    num_xs = [3, 3, 1, 3, 2, 2, 2, 1]
    num_ys = [3, 1, 3, 2, 3, 2, 1, 2]

    total_windows = []
    for num_x, num_y in zip(num_xs, num_ys):
        inter_x = min(int(n_bins / num_x), n_bins - 1)
        inter_y = min(int(n_bins / num_y), n_bins - 1)
        
        for i in range(num_x):
            for j in range(num_y):
                start_x = i*inter_x
                start_y = j*inter_y
                end_x = min(start_x + inter_x, n_bins - 1)
                end_y = min(start_y + inter_y, n_bins - 1)
                total_windows.append([start_x, start_y, end_x, end_y])
    
    start_x, start_y, end_x, end_y = random.choice(total_windows)
else:
    inter = int(n_bins / 3)
    start_x = random.randint(0, inter * 2)
    start_y = random.randint(0, inter * 2)
    rect_w, rect_h = random.randint(inter, n_bins - 1), random.randint(inter, n_bins - 1)
    end_x, end_y = min(start_x + rect_w, n_bins - 1), min(start_y + rect_h, n_bins - 1)

spatial_window_prompt = [start_x, start_y, end_x, end_y]

\end{lstlisting}

\subsection{Table Recognition}

Given a table image, we resize it to 1,024$\times$1,024 pixels. The \pointsdecoder, utilizing the feature vector from the Image Encoder, simultaneously generates pure HTML tags with structural cell point sequences in the same sequence representing the table's logical and physical structures. These structural cell point sequences serve as start-prompting input for the \contentdecoder, which extracts table cell contents in parallel. The final output combines pure HTML tags with cell contents, forming complete HTML sequences faithfully representing the table's structure and content.


\mypara{Datasets.} 
Since our model predicts both the logical structure of tables with cell bounding box central points and cell content, datasets lacking cell content and corresponding bounding box annotations, such as TABLE2LATEX-450K~\cite{deng2019challenges}, TableBank~\cite{li2020tablebank}, UNLV~\cite{shahab2010open}, IC19B2H~\cite{gao2019icdar}, WTW~\cite{long2021parsing} and TUCD~\cite{raja2021visual}, are not suitable for our approach. Similarly, datasets like ICDAR2013Table~\cite{gobel2013icdar}, SciTSR~\cite{chi2019complicated}, and PubTables-1M~\cite{smock2022pubtables}, which provide cell content and content box annotations, employ metrics based on box representations that are incompatible with our point-based format. Consequently, PubTabNet (PTN)~\cite{EDD} and FinTabNet (FTN)~\cite{GTE} are selected for our model evaluation.

\begin{figure}[htbp]
  \centering \includegraphics[width=0.96\linewidth]{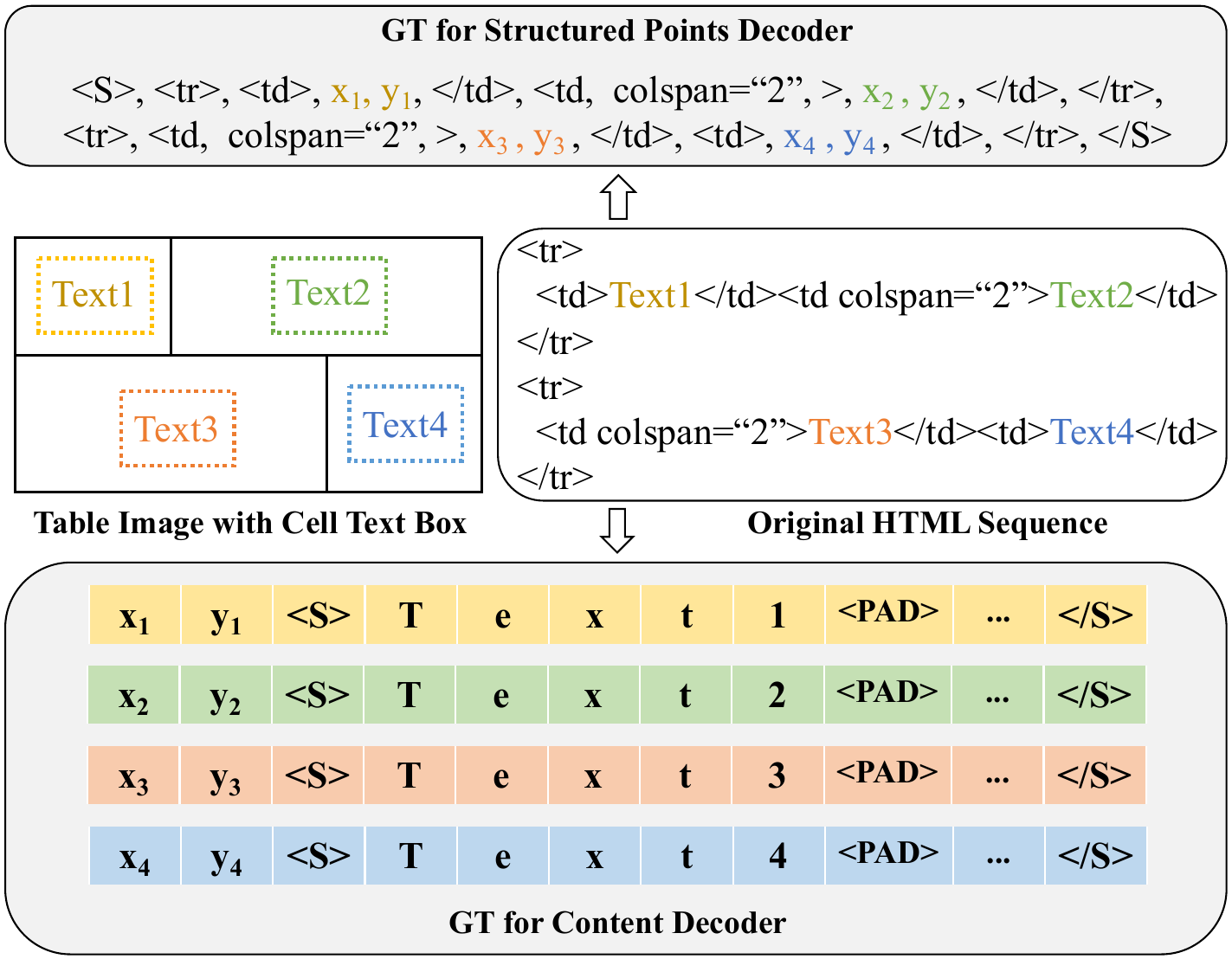}
   \captionsetup{width=0.96\linewidth}
   \caption{\textbf{An Example of building training GTs for table recognition task.} We use the center points of each cell text box to build GTs for the \pointsdecoder and the \contentdecoder. If the cell is empty text, the corresponding points in the GTs are left empty as well. }
   \label{fig:gt_table_stru}
\end{figure}

\mypara{GT Generation.} The ground truth pure HTML tags of tables are tokenized into structural tokens. Following the previous works~\cite{TableMaster, VAST}, we use the merged labels to represent a non-spanning cell to reduce the length of the HTML tags. Specifically, we use 
\emph{\textless td\textgreater\textless/td\textgreater} and \emph{\textless td\textgreater[]\textless/td\textgreater} to denote empty cells and non-empty cells, respectively. For a cell spanning multiple rows or columns, the original HTML tags are broken into four tokens: \emph{\textless td}, \emph{colspan=``n"} or \emph{rowspan=``n"}, \emph{\textgreater}, and \emph{\textless/td\textgreater}. We use the first token \emph{\textless td} to represent a spanning cell. In addition, four special symbol categories need to be added: \emph{\textless S\textgreater}, \emph{\textless /S\textgreater}, \emph{\textless PAD\textgreater}, and \emph{\textless UNK\textgreater}, which represents the beginning of a sequence, the end of a sequence, padding symbols, and unknown characters, respectively. For building the GT of \pointsdecoder, we insert center points of each cell text box to corresponding HTML tags. For building the GT of \contentdecoder, we combine each cell text with corresponding center points as a whole sequence where center points can be viewed as a start-prompting input for recognizing text, and each cell text is tokenized at the character level. An example of building a training sequence GT for the \pointsdecoder and the \contentdecoder in the table recognition task is illustrated in~\cref{fig:gt_table_stru}.

\begin{figure*}[htbp]
  \centering \includegraphics[width=0.95\linewidth]{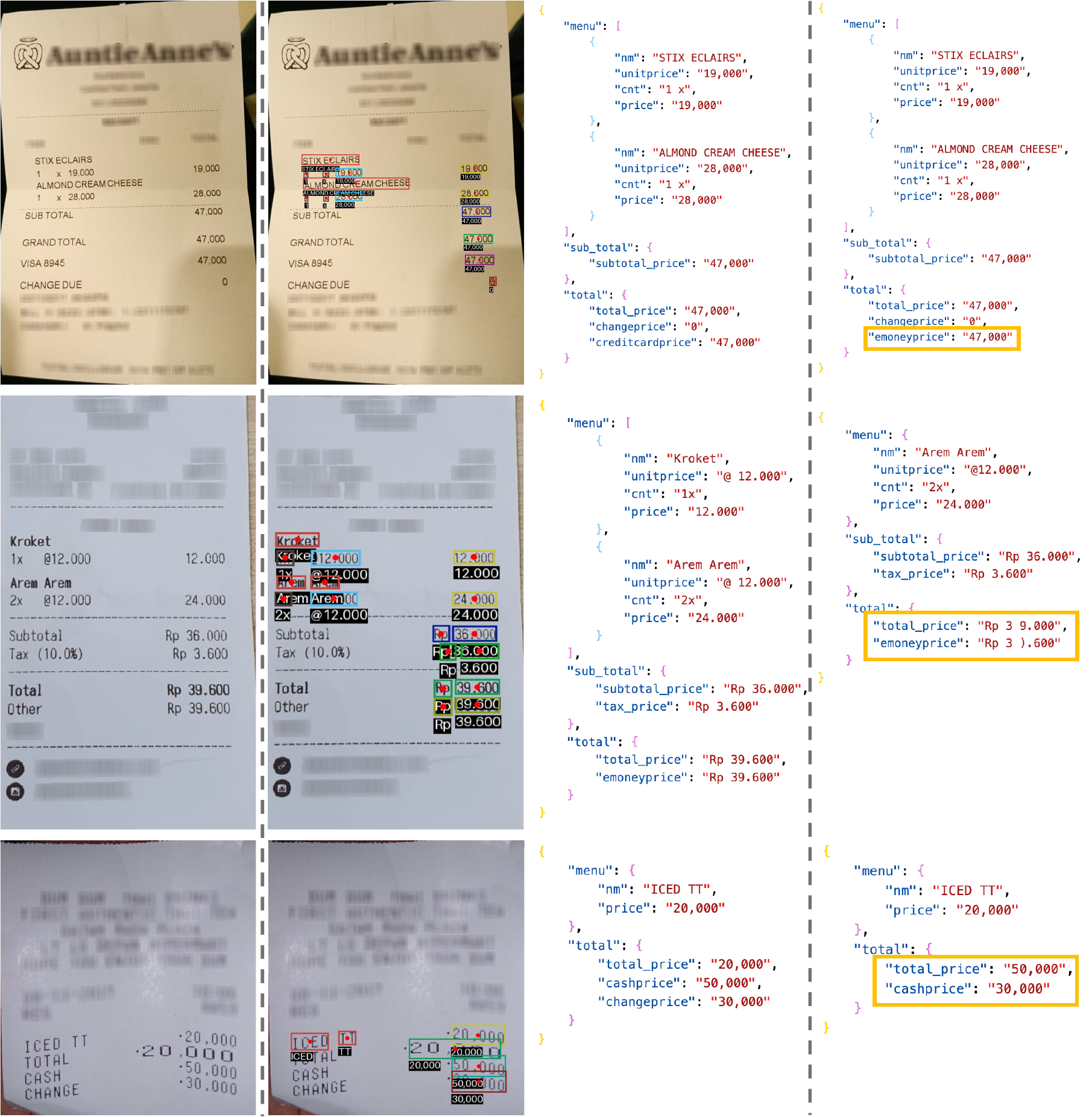}
   \captionsetup{width=0.95\linewidth}
   \caption{\textbf{A comparative analysis of partial results obtained from \ourmodel and Donut on CORD.} The first column depicts the original image, while columns 2 and 3 illustrate our detection results and the corresponding formatted output, respectively. Column 4 showcases the Donut's formatted output. Notably, our model demonstrates superior performance in entity extraction. }
   \label{fig:vs_donut_on_cord}
\end{figure*}

\begin{figure*}[htbp]
  \centering \includegraphics[width=0.95\linewidth]{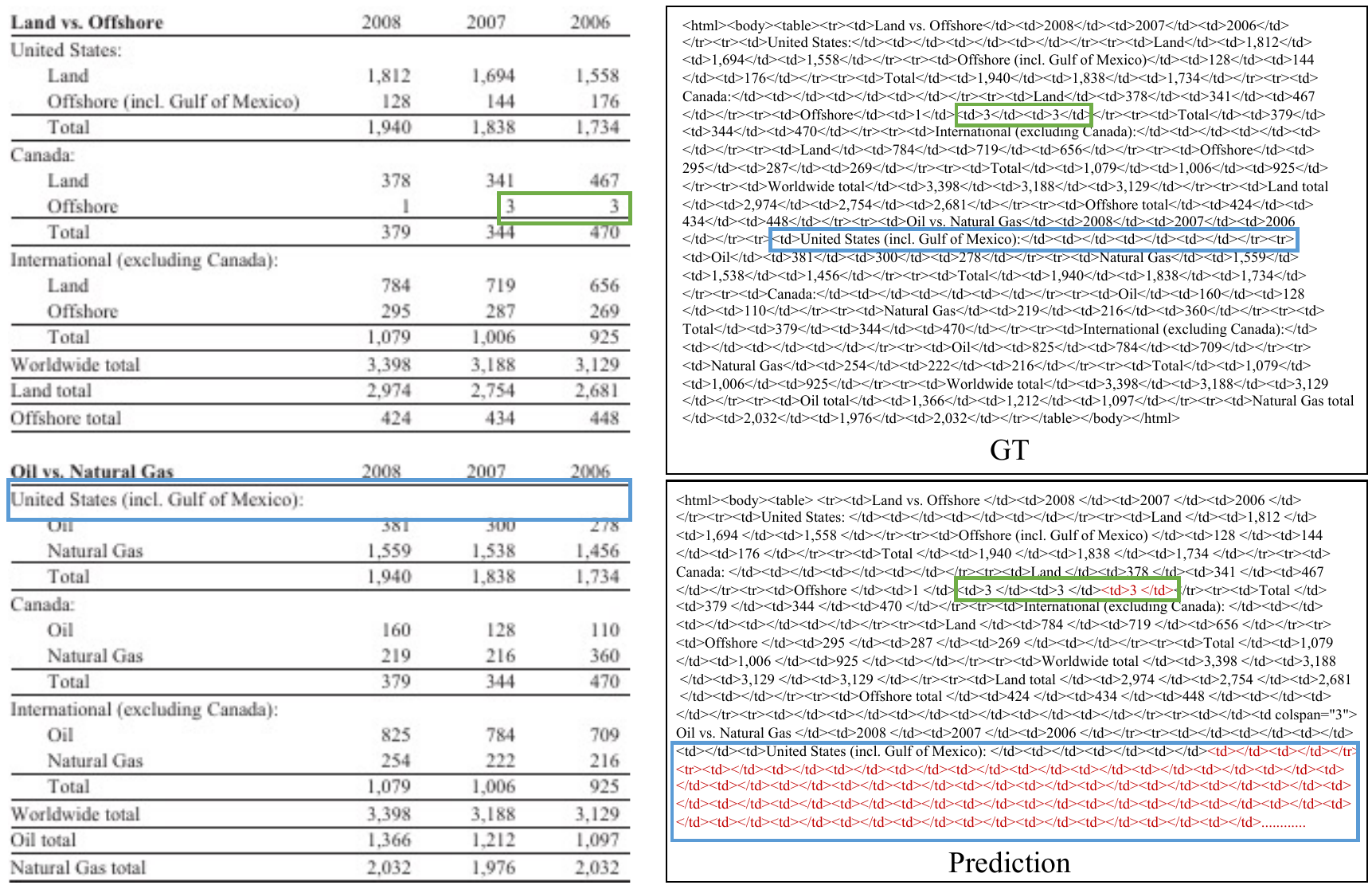}
   \captionsetup{width=0.95\linewidth}
   \caption{\textbf{Illustrative failure case of Donut in table recognition task.} Red text means error predictions. For readability, we only highlight two errors in this example. Due to the lack of point location information, Donut has an attention drift problem, resulting in the prediction of repeated tokens and leading to a high probability of error accumulation in long-sequence scenarios. (The figure is best viewed in color.)}
   \label{fig:donut_table_failure_case}
\end{figure*}

\section{Comparisons with Donut on KIE Task}
As shown in ~\cref{fig:vs_donut_on_cord}, \ourmodel can achieve entity extraction while predicting the location of each entity word.
However, Donut only predicts the structured sequence for entity extraction without any localization ability.
Thus, the absence of direct region supervision during both training and prediction stages often leads to inferior results for entities of same values (Row 1), repeated entities (Row 2) or entities with explicit trigger names (Row 3).

\section{Training Donut on Table Recognition Task}
We fine-tuned the OCR-free end-to-end model Donut~\cite{kim2022donut} for table recognition on FinTabNet dataset. The ground truth sequence utilized combined HTML tags with table cell text, and we use different training hyper-parameters for adequate verification, as shown in \cref{tab:donut_abla}. Due to GPU memory limitations, we constrained the decoder's max length in Donut to 4,000. Note that the original HTML sequence max lengths for PubTabNet and FinTabNet are 8,722 and 8,035, respectively. 
For long sequence prediction tasks such as table recognition, training an end-to-end model like Donut with combined HTML stages, including cell text, is non-trivial.
There is a high probability of error accumulation and attention drift in long-sequence scenarios leading to the inferior performance of Donut for table recognition. An illustrative example of a failure case for Donut in table recognition task is shown in~\cref{fig:donut_table_failure_case}. 
Specifically, due to the lack of region supervision, the end-to-end model Donut has demonstrated an attention drift problem, resulting in the prediction of repeated tokens and leading to a high probability of error accumulation in long-sequence scenarios. In contrast, \ourmodel decomposes the location-aware structured points sequence and cell text recognition generation, alleviating the issues of attention drift and error accumulation.

\begin{table}[htbp]
\centering
   \begin{adjustbox}{max width=0.47\textwidth}
   \begin{threeparttable}
\begin{tabular}{lccccc}
\toprule
 
  Methods                            & LR      & Epoch      & S-TEDS & TEDS  \\
 \midrule

\multirow{5}{*}{Donut~\cite{kim2022donut} }                                                   & 3e-5            & 20               &  22.2  & 17.2  \\
                                                 & 3e-5            & 40               &  26.2  & 20.0  \\
                                                   & 1e-4            & 40               &  30.7  & 29.1  \\
                                                    & 1e-3            & 40               &  41.7  & 40.5  \\
                                                   & 1e-3            & 100               &   41.9 & 41.2  \\
 \midrule
\multirow{1}{*}{\ourmodel (ours)}  &   -          & -             & \textbf{91.55}  & \textbf{89.75} \\

\bottomrule
\end{tabular}
   \end{threeparttable}
   \end{adjustbox}
\caption{\textbf{Comparisons of different training hyper-parameters of Donut on FinTabNet datasets.} LR is short for learning rate.}
\label{tab:donut_abla}
\vspace{-5mm}
\end{table}

\begin{figure}[htbp]
  \centering \includegraphics[width=0.98\linewidth]{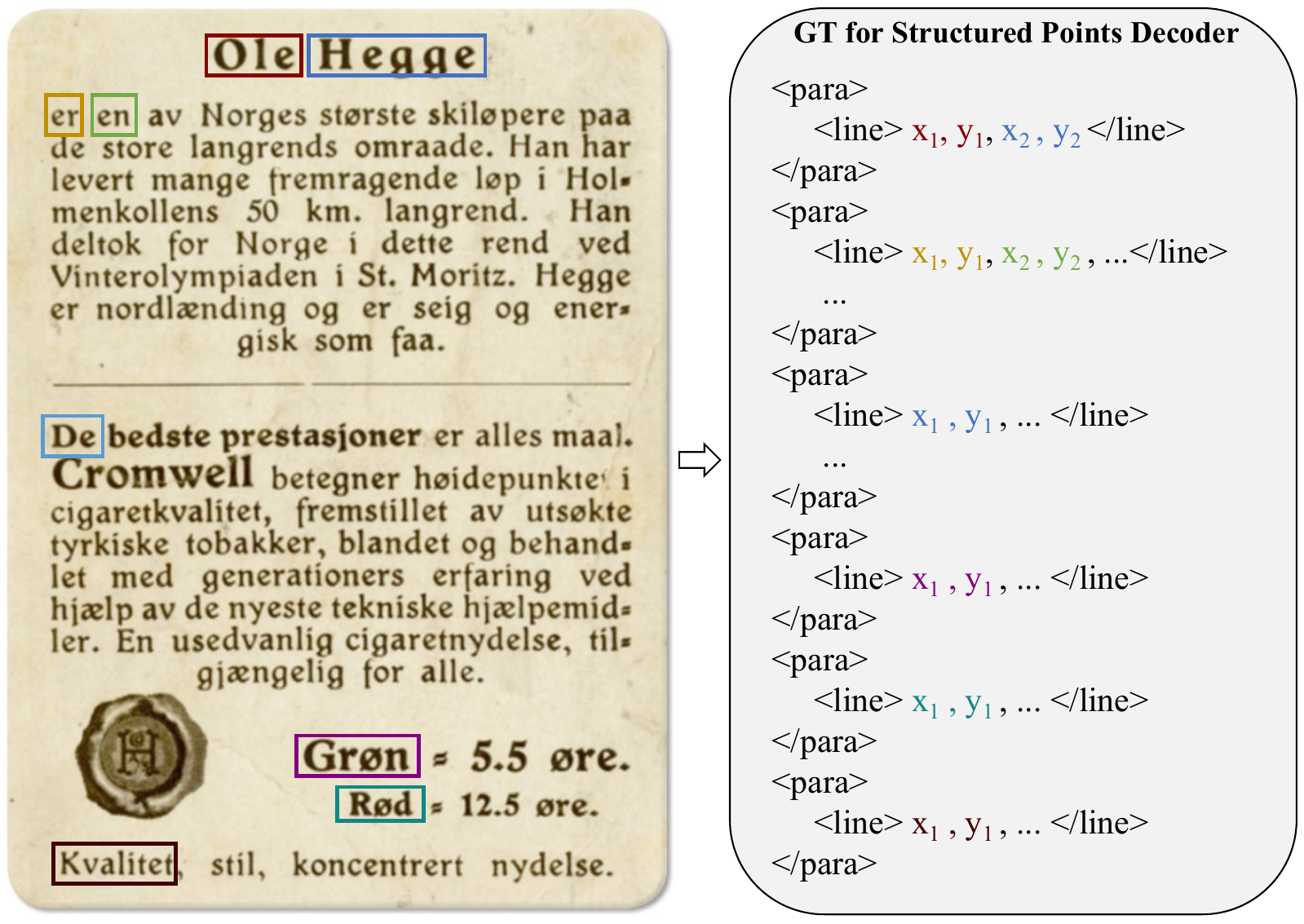}
   \captionsetup{width=0.98\linewidth}
   \caption{\textbf{An Example of building training GTs for hierarchical text detection task.} }
   \label{fig:gt_layout_stru}
\end{figure}

\section{Generalization to Hierarchical Text Detection Task}
\begin{figure*}[htbp]
  \centering \includegraphics[width=0.70\linewidth]{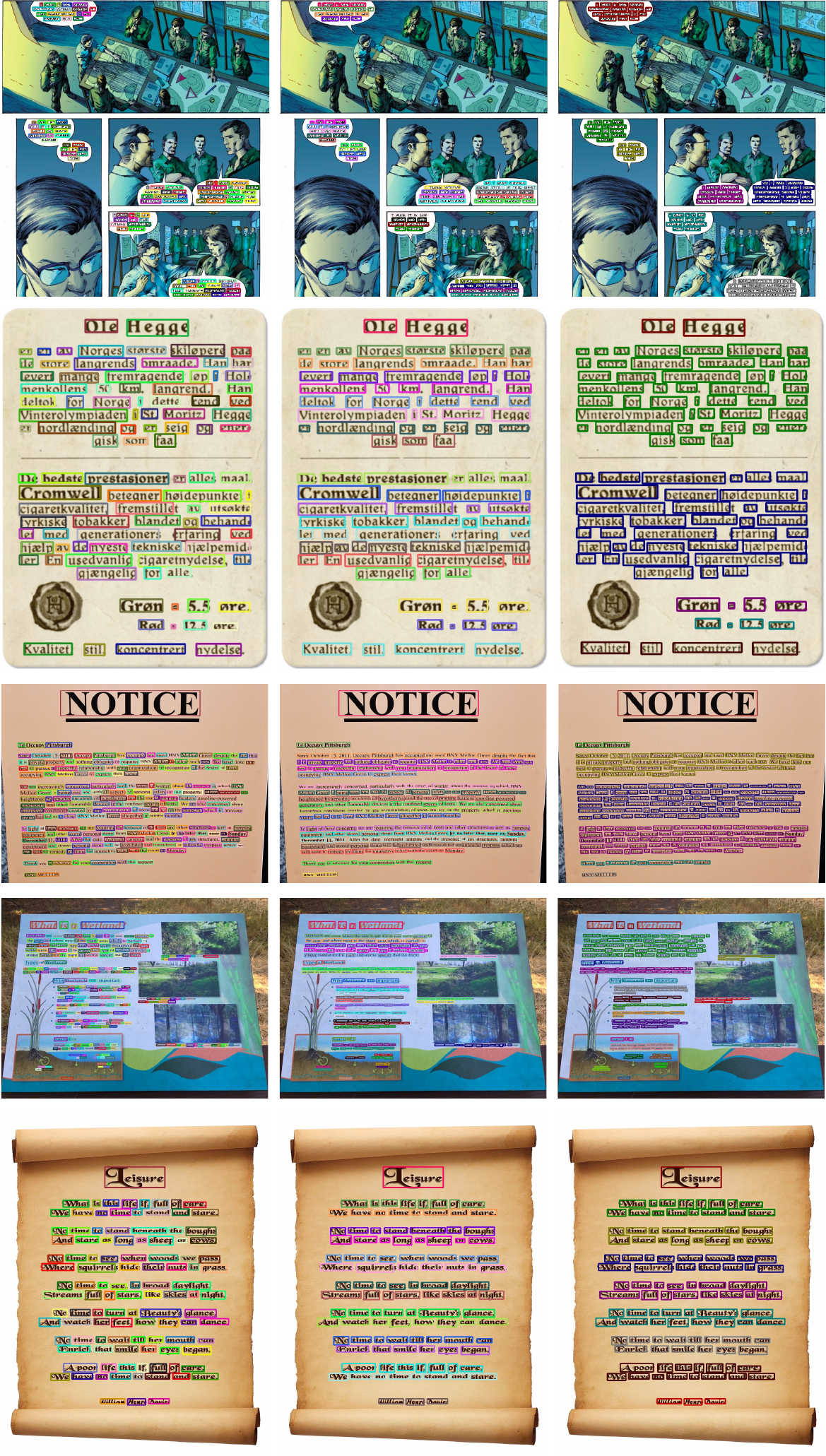}
   \captionsetup{width=0.98\linewidth}
   \caption{\textbf{Visualization results of hierarchical text detection.} Columns 1-3 represent the detection results for word, line, and paragraph levels, respectively. Text instances belonging to the same hierarchical level are enclosed within rectangles of the same color. (The figure is best viewed in color.)}
   \label{fig:vis_hiertext}
\end{figure*}

Thanks to the flexible expression of structured sequence in \ourmodel, it is convenient for us to extend it to other OCR-related tasks, such as hierarchical text detection, which aims to group the text in the image into three levels, namely word, line, and paragraph, based on spatial position and semantic relationship. Previous methods~\cite{long2022towards} mainly achieved hierarchical results by clustering based on similarity. In our approach, we distinguish the text belonging to different hierarchical intervals by simply inserting \emph{\textless LINE\textgreater} and \emph{\textless PARA\textgreater} structural tags into the sequence of text center points, as shown in~\cref{fig:gt_layout_stru}. The experiments are mainly conducted on the HierText dataset~\cite{long2022towards}, which consists of 8,281 training images, 1,724 validation images, and 1,634 test images. We train the model on the training set and evaluate on the validation set. Partial visualization results are shown in~\cref{fig:vis_hiertext}. 
Without any task-specific architectural designs, our model achieves promising results, demonstrating its strong generalization ability. 

\begin{figure*}[htbp]
  \centering \includegraphics[width=0.76\linewidth]{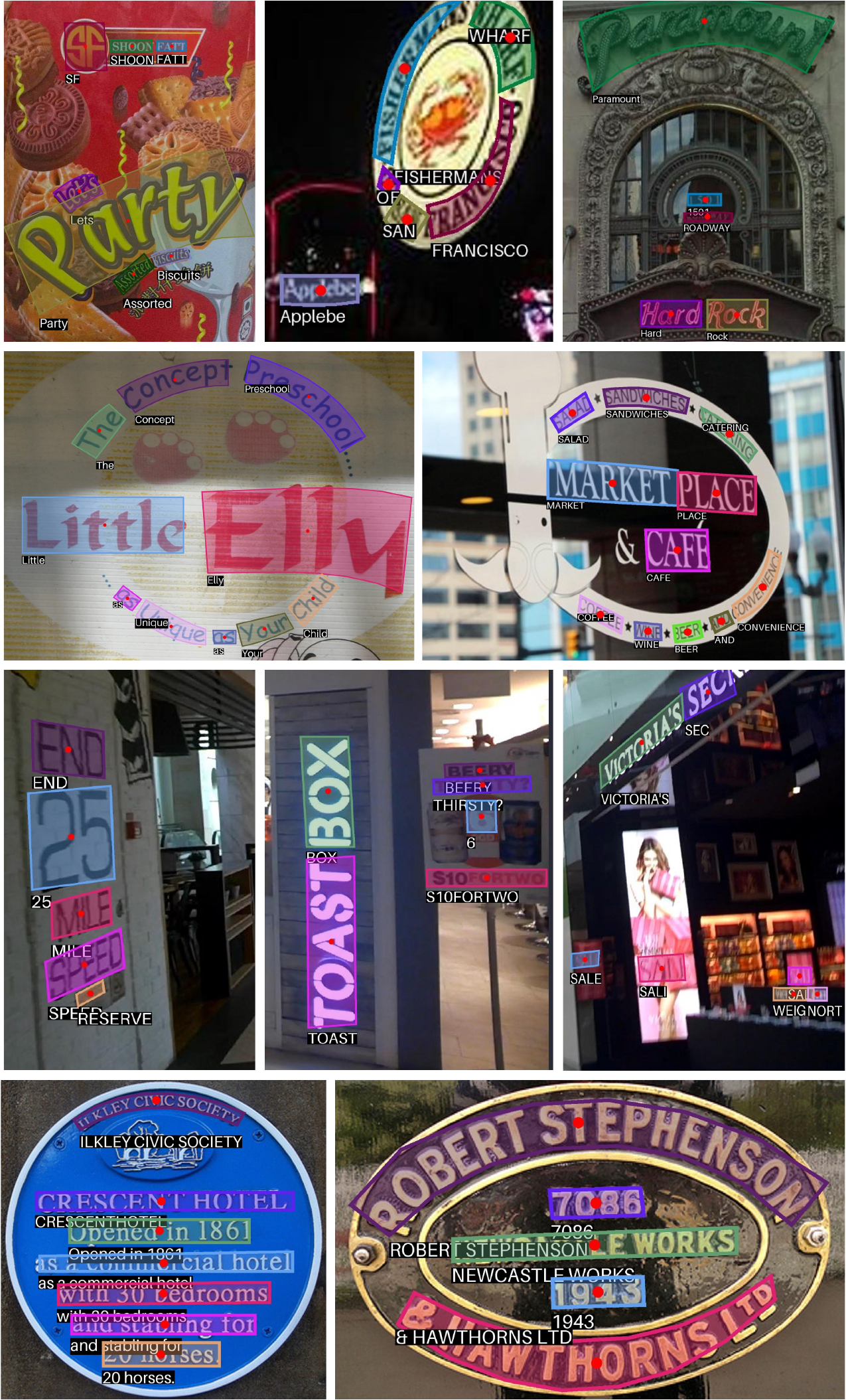}
   \captionsetup{width=0.98\linewidth}
   \caption{\textbf{Visualization results of text spotting.} Rows 1-2 depict the visual results on the Total-Text dataset, while rows 3 and 4 respectively illustrate the visual results on the ICDAR 2015 and CTW1500 datasets. (The figure is best viewed in color.)}
   \label{fig:vis_spotting_more}
\end{figure*}

\begin{figure*}[htbp]
  \centering \includegraphics[width=0.95\linewidth]{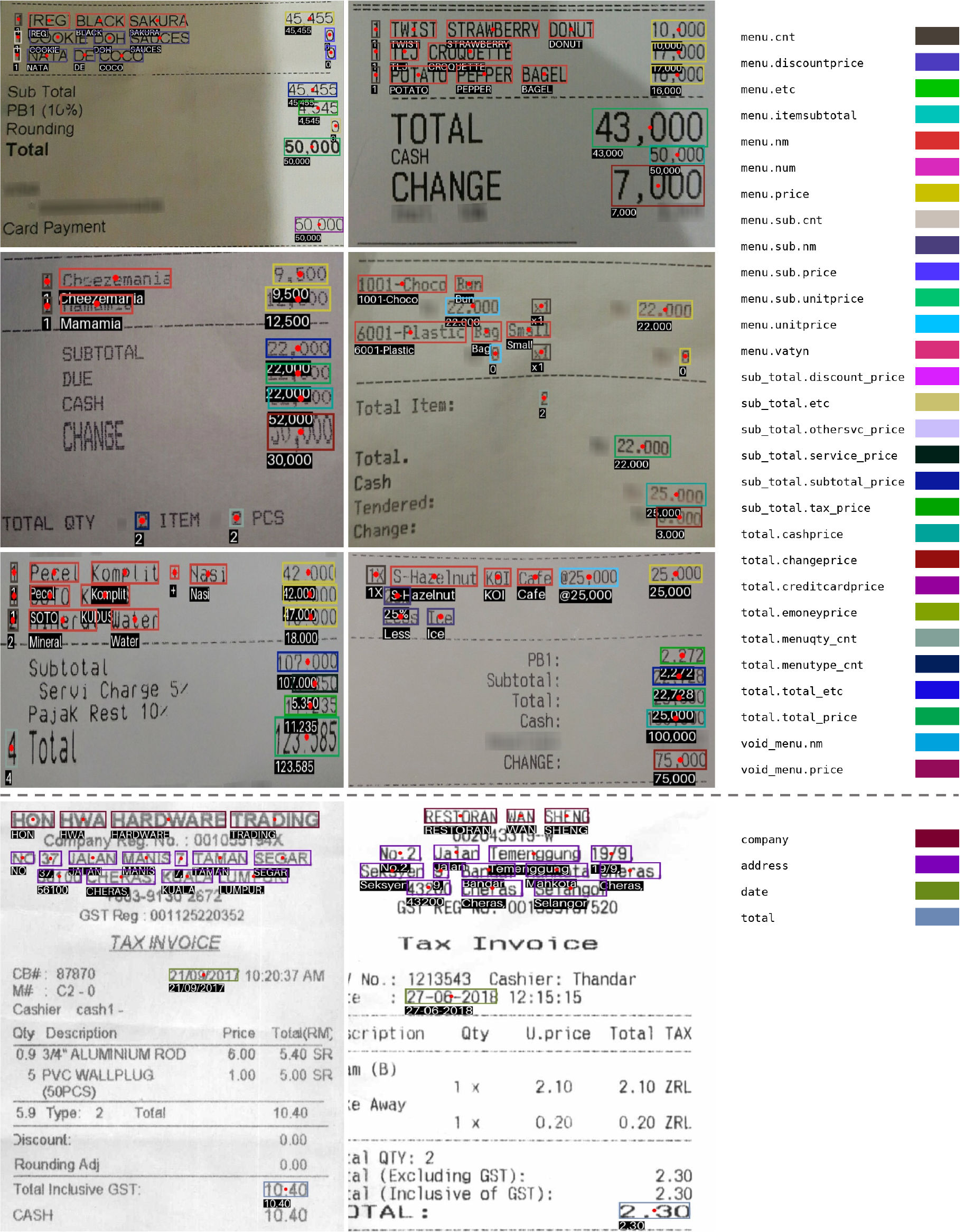}
   \captionsetup{width=0.98\linewidth}
   \caption{\textbf{Visualization results of key information extraction.} Rows 1-3 and row 4 demonstrate the visual results on the CORD and SROIE datasets respectively. In order to differentiate entities of different categories, we employ rectangles of varying colors. The correspondence between colors and categories can be seen in the legend on the right side. (The figure is best viewed in color.)}
   \label{fig:vis_kie_more}
\end{figure*}

\begin{figure*}[htbp]
  \centering \includegraphics[width=0.98\linewidth]{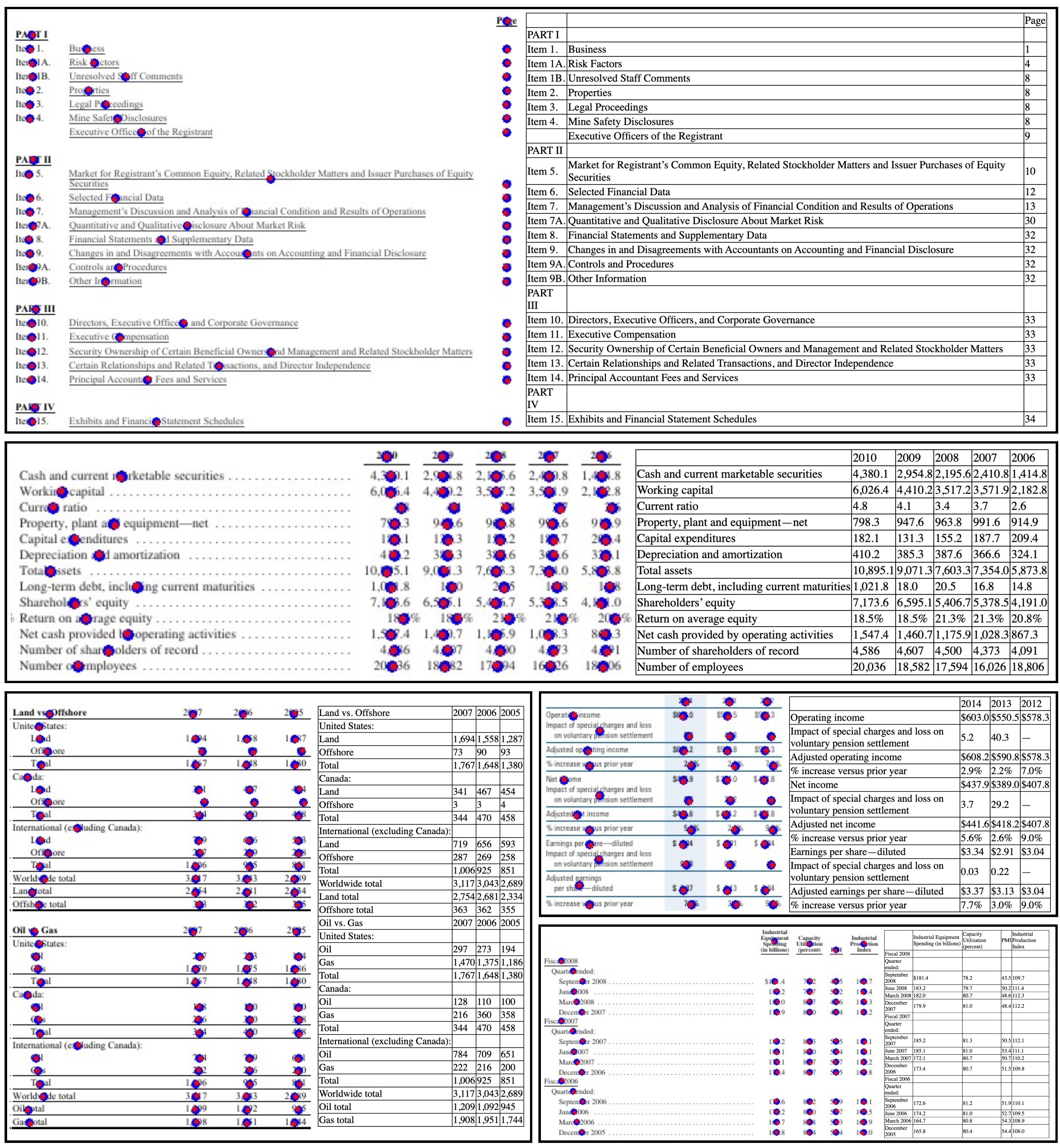}
   \captionsetup{width=0.98\linewidth}
   \caption{\textbf{Visualization results of table recognition.} We present point locations and a rendered table with an additional border for readability based on the prediction sequence in each group. Blue points and red points denote the GT and predicted points respectively. (The figure is best viewed in color.)}
   \label{fig:vis_table_more}
\end{figure*}


\section{More Visualizations}
~\cref{fig:vis_spotting_more}, ~\cref{fig:vis_kie_more}, and~\cref{fig:vis_table_more} are more qualitative results of text spotting, key information extraction, and table recognition, respectively.

{
    \small
    \bibliographystyle{ieeenat_fullname}
    \bibliography{main}
}
